\documentclass[runningheads]{llncs}

\usepackage[mobile]{eccv}

\usepackage{eccvabbrv}

\usepackage[T1]{fontenc}
\usepackage[breaklinks,colorlinks,pagebackref]{hyperref}
\usepackage{graphicx}
\usepackage{booktabs}
\usepackage{amsmath}

\newcommand{\keypoint}[1]{\vspace{0.1cm}\noindent\textbf{#1}}

\def\modelName{\textsc{UnSCAR}\xspace}

\usepackage{ulem}
\usepackage{booktabs} 
\usepackage{tabularx}
\usepackage{ragged2e}

\usepackage{xcolor}

\usepackage{svg}
\usepackage{colortbl}
\usepackage{microtype}

\usepackage{pifont}
\newcommand{\cmark}{\ding{51}}%
\newcommand{\xmark}{\ding{55}}%

\usepackage{multirow} 
\usepackage[table,dvipsnames]{xcolor} 

\usepackage{amssymb}
\usepackage[inline]{enumitem}

\newcommand{\greencheck}{{\color{Green}\cmark}\xspace}

\newcommand{\yellowcheck}{{\color{YellowOrange}(\cmark)}\xspace}
\newcommand{\redcheck}{{\color{red}\xmark}\xspace}

\definecolor{darkorchid}{rgb}{0.6,0.196,0.8}

\definecolor{yujieColor}{rgb}{0,0.4,0}

\usepackage[accsupp]{axessibility}  

\makeatletter
\renewcommand{\paragraph}{%
  \@startsection{paragraph}{4}%
  {\z@}{0.2ex \@plus 0.3ex \@minus .2ex}{-1em}%
  {\normalfont\normalsize\bfseries}%
}
\makeatother

\usepackage{orcidlink}

\begin{document}

\title{\textsc{UnSCAR}: \underline{Un}iversal, \underline{S}calable, \underline{C}ontrollable,\\ and \underline{A}daptable Image \underline{R}estoration} 

\titlerunning{\textbf{Un}iversal, \textbf{S}calable, \textbf{C}ontrollable and \textbf{A}daptable Image \textbf{R}estoration}

\author{Debabrata Mandal \inst{1} \hspace{0.05cm}
Soumitri Chattopadhyay\inst{2} \hspace{0.05cm}
Yujie Wang\inst{1} \hspace{0.05cm} \\Marc Niethammer\inst{2} \hspace{0.05cm} Praneeth Chakravarthula\inst{1}}

\authorrunning{D.~Mandal et al.}

\institute{University of North Carolina at Chapel Hill \and
University of California, San Diego\\}

\maketitle

\begin{abstract}

Universal image restoration aims to recover clean images from arbitrary real-world degradations using a single inference model. 
Despite significant progress, existing all-in-one restoration networks do not scale to multiple degradations.
As the number of degradations increases, training becomes unstable, models grow excessively large, and performance drops across both seen and unseen domains. 
In this work, we show that scaling universal restoration is fundamentally limited by interference across degradations during joint learning, leading to \textit{catastrophic task forgetting}.
To address this challenge, we introduce a unified inference pipeline with a multi-branch mixture-of-experts architecture that decomposes restoration knowledge across specialized task-adaptable experts.
Our approach enables \textit{scalable} learning (over sixteen degradations), \textit{adapts and generalizes} robustly to unseen domains, and supports \textit{user-controllable restoration} across degradations. 
Beyond achieving superior performance across benchmarks, this work establishes a new design paradigm for scalable and controllable universal image restoration.

\end{abstract}

\begin{figure}
\centering
\vspace{-3mm}
\includegraphics[width=\textwidth]{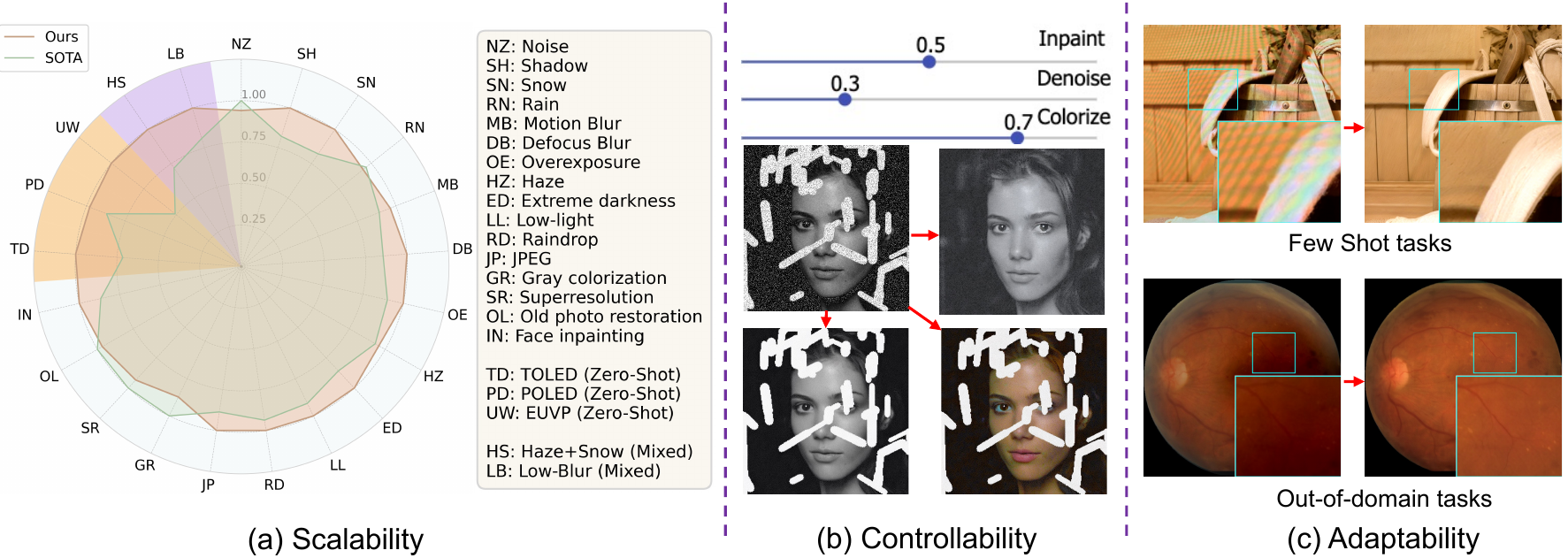}
\caption{
We introduce \modelName, a universal image restoration model that can handle $2\times$ more degradations than previously explored, achieving state-of-the-art performance across multiple benchmarks. Our framework also enables user-controllable restoration via degradation-control sliders and supports few-shot adaptation (down to one-shot) with strong out-of-distribution generalization on challenging medical imaging datasets.}
\label{fig:teaser}
\end{figure}

\section{Introduction}
\label{sec:intro}

Image restoration has progressed from heuristic pixel-level algorithms to powerful black-box deep neural network models \cite{chen2022nafnet, zamir2022restormer, dong2024dehazedct, song2023dehazeformer, restorevar, potlapalli2024promptir, mandal2025unicorn, defusion, gandelsman2019double, guo2024onerestore, jiang2023autodir, luo2023controlling, rajagopalan2024awracle,li2025diffusion}. Recent approaches demonstrate that a single architecture can address multiple restoration tasks, simplifying model design, storage, and deployment across different ill-posed imaging problems \cite{zamir2022restormer, potlapalli2024promptir, gandelsman2019double,chen2022nafnet}. 
However, these unified architectures typically require \textit{separate training for each degradation type} \cite{zamir2022restormer, potlapalli2024promptir}. 
As a result, they struggle with mixed or unseen degradations, where models must first identify degradation before performing restoration  \cite{jiang2023autodir, luo2023controlling}.

\begin{table}[!t]
    \setlength{\tabcolsep}{0.2em} 
    \centering
    \footnotesize 
    \caption{
        Comparison with prior image restoration works. 
        Factors are denoted as fully met~\greencheck, partially met~\yellowcheck, or unmet~\redcheck.
    }
    \vspace{-0.2cm}
    
    \newcolumntype{Y}{>{\centering\arraybackslash}X}
    
    \resizebox{\textwidth}{!}{
    \setlength{\tabcolsep}{4pt}
    \setlength{\extrarowheight}{2pt}
    \begin{tabular}{ccccccccc}
    \toprule
    \textbf{Method} & 
    \textbf{Venue} & 
    \scriptsize \textbf{Unified Model} & 
    \scriptsize \textbf{Mixed Restoration} & 
    \scriptsize \textbf{Automatic} & 
    \scriptsize \textbf{Scalability} & 
    \scriptsize \textbf{Adaptability} & 
    \scriptsize \textbf{Control Modality} & 
    \scriptsize \textbf{\#Primary Tasks} \\
    \midrule
    
    Restormer~\cite{zamir2022restormer} & 
    CVPR`22 & 
    \redcheck & 
    \redcheck & 
    \redcheck & 
    \redcheck & 
    \yellowcheck & 
    \textemdash & 
    1 \\
    
    AutoDIR~\cite{jiang2023autodir} & 
    ECCV`24 & 
    \greencheck & 
    \yellowcheck & 
    \yellowcheck & 
    \redcheck & 
    \redcheck & 
    \textemdash & 
    4 \\
    
    DA-CLIP~\cite{luo2023controlling} & 
    ICLR`24 & 
    \redcheck & 
    \redcheck & 
    \greencheck & 
    \redcheck & 
    \redcheck & 
    \textemdash & 
    10 \\

    OneRestore~\cite{guo2024onerestore} & 
    ECCV`24 & 
    \greencheck & 
    \greencheck & 
    \yellowcheck & 
    \yellowcheck & 
    \redcheck & 
    Limited & 
    3 \\

    MPerceiver~\cite{mperceiver} & 
    CVPR`24 & 
    \greencheck & 
    \greencheck & 
    \greencheck & 
    \redcheck & 
    \yellowcheck & 
    \textemdash & 
    8 \\

    CAPTNet~\cite{captnet} & 
    TCVST`24 & 
    \greencheck & 
    \redcheck & 
    \redcheck & 
    \redcheck & 
    \redcheck & 
    \textemdash & 
    4 \\
    
    RestoreVAR~\cite{restorevar} & 
    ICLR`26 & 
    \greencheck & 
    \greencheck & 
    \yellowcheck & 
    \redcheck & 
    \redcheck & 
    \textemdash & 
    4 \\

    Defusion~\cite{defusion} & 
    CVPR`25 & 
    \greencheck & 
    \yellowcheck & 
    \greencheck & 
    \greencheck & 
    \redcheck & 
    \textemdash & 
    8 \\

    UniRes~\cite{unires} & 
    ICCV`25 & 
    \greencheck & 
    \greencheck & 
    \greencheck & 
    \yellowcheck & 
    \yellowcheck & 
    \textemdash & 
    4 \\

    UARE~\cite{uare} & 
    arXiv`25 & 
    \greencheck & 
    \greencheck & 
    \greencheck & 
    \redcheck & 
    \yellowcheck & 
    Text & 
    5 \\

    FluxIR~\cite{fluxir} & 
    CVPR`25 & 
    \greencheck & 
    \redcheck & 
    \yellowcheck & 
    \yellowcheck & 
    \redcheck & 
    Text & 
    2 \\

    UniCoRN~\cite{mandal2025unicorn} & 
    WACV`26 & 
    \greencheck & 
    \greencheck & 
    \yellowcheck & 
    \greencheck & 
    \yellowcheck & 
    Text & 
    6 \\
    
    \midrule
    
    \rowcolor{cyan!20} \textsc{\textbf{UnSCAR}} \textit{\textbf{(Ours)}} & 
    \textemdash & 
    \greencheck & 
    \greencheck & 
    \greencheck & 
    \greencheck & 
    \greencheck & 
    Sliders & 
    \textbf{16} \\
    
    \bottomrule
    \end{tabular}
    }
    \label{tab:related_work}
    \vspace{-0.1cm}
\end{table}

Recent universal or all-in-one restoration (AIR) methods attempt to address this challenge by jointly learning multiple degradations within a single framework. This allows for restoration under mixed or unknown conditions \cite{mandal2025unicorn, luo2023controlling, mperceiver, rajagopalan2024awracle,restorevar}.
However, AIR models trained across multiple degradation removal tasks often suffer from catastrophic forgetting \cite{wu2025re, kemker2018measuring}.
This leads to a performance drop as the number and diversity of degradation types increase. 
Furthermore, generative restoration models trained on natural scenes often fail on out-of-distribution data, such as surgical \cite{laparoscopic}, medical \cite{fundus}, or satellite imagery \cite{akhatova2025stable}.
When guided by standard control modalities (e.g., text prompts) \cite{mperceiver,uare,fluxir}, these models frequently struggle due to limited domain-specific training data.
As a result, current AIR approaches remain \textit{restricted} to training data domains and lack mechanisms for \textit{forgetting-free adaptation}.
Additionally, most existing AIR methods focus on blind restoration, providing limited control over the image recovery process.

In this work, we address these challenges by introducing \textbf{\modelName}, a scalable and controllable universal image restoration framework capable of handling multiple degradations (more than 16 demonstrated), significantly extending the scope of prior approaches \cite{mandal2025unicorn, luo2023controlling, mperceiver} (see Tab.~\ref{tab:related_work}).
Our method builds upon the UniCoRN architecture \cite{mandal2025unicorn}, which introduced low-level visual cues and mixture-of-experts (MoE) adapters for generative restoration.
We extend this design with several architectural improvements aimed at improving scalability and robustness:
\textbf{\textit{(i)}} unified encoding of richer low-level visual priors, \textbf{\textit{(ii)}} task-aware degradation--content disentangled embeddings, \textbf{\textit{(iii)}} bi-directional feedback between control and generative branches, and \textbf{\textit{(iv)}} novel residual-attention MoE blocks that reduce cross-task interference.
Extensive experiments across single, synthetically mixed, and complex degradation benchmarks demonstrate the robustness, scalability and universality of the proposed \modelName architecture.

Beyond scalability, we introduce \textit{fine-grained controllability} into the restoration process. 
Inspired by controllable generative models \cite{rombach2022high, midjourney, flux} and concept editing methods \cite{conceptsliders, conceptediting}, we propose \textbf{\textit{(v)}} a degradation control slider mechanism that allows users to modify restoration behavior. 
As illustrated in Fig.~\ref{fig:teaser}, the slider softly reweights expert activations, overriding the default MoE-based restoration strategy to enable interpretable and flexible degradation manipulation.
Finally, we demonstrate that \modelName can \textit{adapt efficiently to unseen restoration} tasks. 
Using \textbf{\textit{(vi)}} a lightweight parameter-efficient fine-tuning (PEFT) approach, our model can be adapted to new restoration tasks in \textit{zero-shot} or \textit{few-shot} settings ($\sim$1–50 samples). 
Our experiments on out-of-distribution medical imaging tasks, including laparoscopic image de-smoking \cite{laparoscopic} and fundus image restoration \cite{fundus}, demonstrate robust generalization and efficient adaptation.



As summarized in Tab.~\ref{tab:related_work}, \modelName constitutes a paradigm shift for scalable and controllable universal image restoration, demonstrating improved flexibility for practical image restoration applications.

\vspace{-0.4cm}
\section{Related works}
\label{sec:related}
\vspace{-0.4cm}


\keypoint{Single-degradation Image Restoration:} 
Early deep-learning approaches for image restoration focused on addressing a \textit{single known degradation}, such as haze, smoke, noise, low-light, or low-contrast. 
Representative task-specific models include those for dehazing \cite{song2023dehazeformer, dong2024dehazedct, li2021you}, denoising \cite{ lehtinen2018noise2noise, fan2019brief}, and deblurring \cite{zhang2022deep, cho2021rethinking}. 
Another line of work proposed unified architectures that can be trained separately for different restoration tasks \cite{gandelsman2019double, zamir2022restormer, wang2022uformer, li2020zero, wang2022uformer, liang2021swinir}. 
These enable end-to-end mapping from degraded images to clean outputs, and remain largely task-agnostic. 
However, they suffer from two fundamental limitations: (i) limited scalability to diverse real-world degradations \cite{mandal2025unicorn}, and (ii) limited availability, or unavailability, of high-quality degradation-specific training data in practice.
Overcoming these limitations, we propose a \textit{unified, degradation-agnostic, scalable and adaptable universal image restoration model} for practical real-world scenarios.

\keypoint{Multi-degradation Image Restoration:} Recent work has focused on simultaneously modeling multiple degradations to reflect real-world images which typically contain several co-existing corruptions \cite{li2022all, luo2023controlling, jiang2023autodir, unires, uare, mandal2025unicorn}. 
These approaches include degradation-specific modules \cite{chen2021pre, valanarasu2022transweather} or embeddings \cite{potlapalli2024promptir, li2022all}, neural architecture search \cite{li2020all}, and degradation-aware prompt-based conditioning \cite{li2023prompt, potlapalli2024promptir}. 
While these methods show improved robustness, they remain limited in scalability and adaptability to diverse degradations \cite{mandal2025unicorn, restorevar, uare}, restricting their practical use -- challenges which we address in this work. 

\keypoint{Conditional Diffusion Models and Controllable Generation:} 
Diffusion models \cite{ho2020denoising, kawar2022denoising, rombach2022high, controlnet} have recently emerged as the de facto paradigm for controllable visual generation across images \cite{rombach2022high}, videos \cite{blattmann2023align}, medical scans \cite{maisiv2}, and 3D objects \cite{poole2022dreamfusion}. 
Popular frameworks include Stable Diffusion \cite{rombach2022high}, ControlNet \cite{controlnet}, DreamBooth \cite{ruiz2023dreambooth}, Flux \cite{flux}, and Wan \cite{wan}.
Conditional diffusion models allow users to control the generation process through signals such as text prompts \cite{rombach2022high}, structural guidance (e.g., edge, depth, color maps) \cite{controlnet, zhao2024unicontrolnet}, or via personalization \cite{ruiz2023dreambooth}. 
Extensions to the ControlNet architecture \cite{zhang2023adding} have further enhanced controllability, such as ControlNet++ \cite{li2024controlnet++} which uses consistency rewards, ControlNet-XS \cite{zavadski2024controlnet} which  provides early feedback injection, UniControlNet \cite{zhao2023uni} for fusing control information, and ControlNeXt \cite{peng2024controlnext} for efficient fine-tuning. 
More recently, low-rank adapters have enabled fine-grained editing in generative models \cite{conceptediting, conceptsliders}. 
Inspired by these advances, our work introduces mechanisms for \textit{adaptability} and \textit{precise control} for real-world image restoration.

\keypoint{Universal Image Restoration via Generative Models:}
Recent research has explored universal image restoration using conditional diffusion models \cite{li2025diffusion, he2024diffusion, zheng2024selective, luo2023controlling}, where the conditioning signal encodes degradation-relevant information. 
For example, DA-CLIP \cite{luo2023controlling} and D4IR \cite{wang2024data} train degradation-aware encoders using contrastive learning \cite{clip} to guide diffusion-based restoration. 
AutoDIR \cite{jiang2023autodir} identifies the dominant degradation before applying a structure-corrected diffusion model for restoration. 
MPerceiver \cite{mperceiver} combines prompt learning and cross-modal feature inversion for stronger visio-linguistic synergy for text-conditioned restoration.
More recent works further extend this paradigm. UniCoRN \cite{mandal2025unicorn} employs low-level visual estimates with ControlNet-style conditioning \cite{controlnet} and MoE modules, RestoreVAR \cite{restorevar} uses visual auto-regressive restoration, UniRes \cite{unires} generates multiple text-driven restorations and combines them at inference time, and UARE \cite{uare} aligns image quality assessment and restoration learning through interleaved data and a progressive schedule for learning single to multiple degradations. 
{\modelName} follows this conditional generative restoration paradigm while introducing several architectural innovations for improving scalability and adaptability across diverse degradation settings.



\vspace{-0.2cm}
\section{Methodology}
\label{sec:method}

\subsection{Preliminaries: Unified Controllable Restoration}\label{sec:pre}

Given a degraded image $\tilde{I}$, an image restoration process typically involves:
(1) identifying present degradations,
(2) balancing their restoration strengths, and
(3) predicting a residual restoration direction $R$ such that $I^*=\tilde{I}+R$.
\textsc{UniCoRN}~\cite{mandal2025unicorn} realizes this paradigm in a latent diffusion framework through conditioning mechanisms and control pathways for generative restoration.
It extracts task-specific low-level visual cues (e.g., edge, haze, noise maps) to identify degradations, 
uses lightweight MoE adapters to weight restoration strengths across the identified degradation types, 
and injects the balanced control signals into the latent space denoiser via a ControlNet-style mechanism.

\paragraph{Limitations of UniCoRN.} While effective, this design faces scalability challenges: 
(1) Degradation-specific cue selection requires manually defined mappings, which become prohibitive as the number of degradations grows. 
(2) Maintaining separate control generation pathways per degradation through ControlNet introduces significant parameter overhead and limits scalability.
(3) UniCoRN employs ultra-lightweight MoE adapters to generate spatially varying weights that balance restoration signals across degradation types, and adopts curriculum learning for mixed degradations. However, as degradation types increase, expert routing becomes increasingly difficult to interpret and control, particularly under limited mixed-degradation training data. These limitations motivate our redesigned architecture for scalable and user-controllable restoration.


\begin{figure}[t]
    \centering
    \begin{subfigure}[b]{0.5\linewidth}
        \centering
        \includegraphics[width=\linewidth]{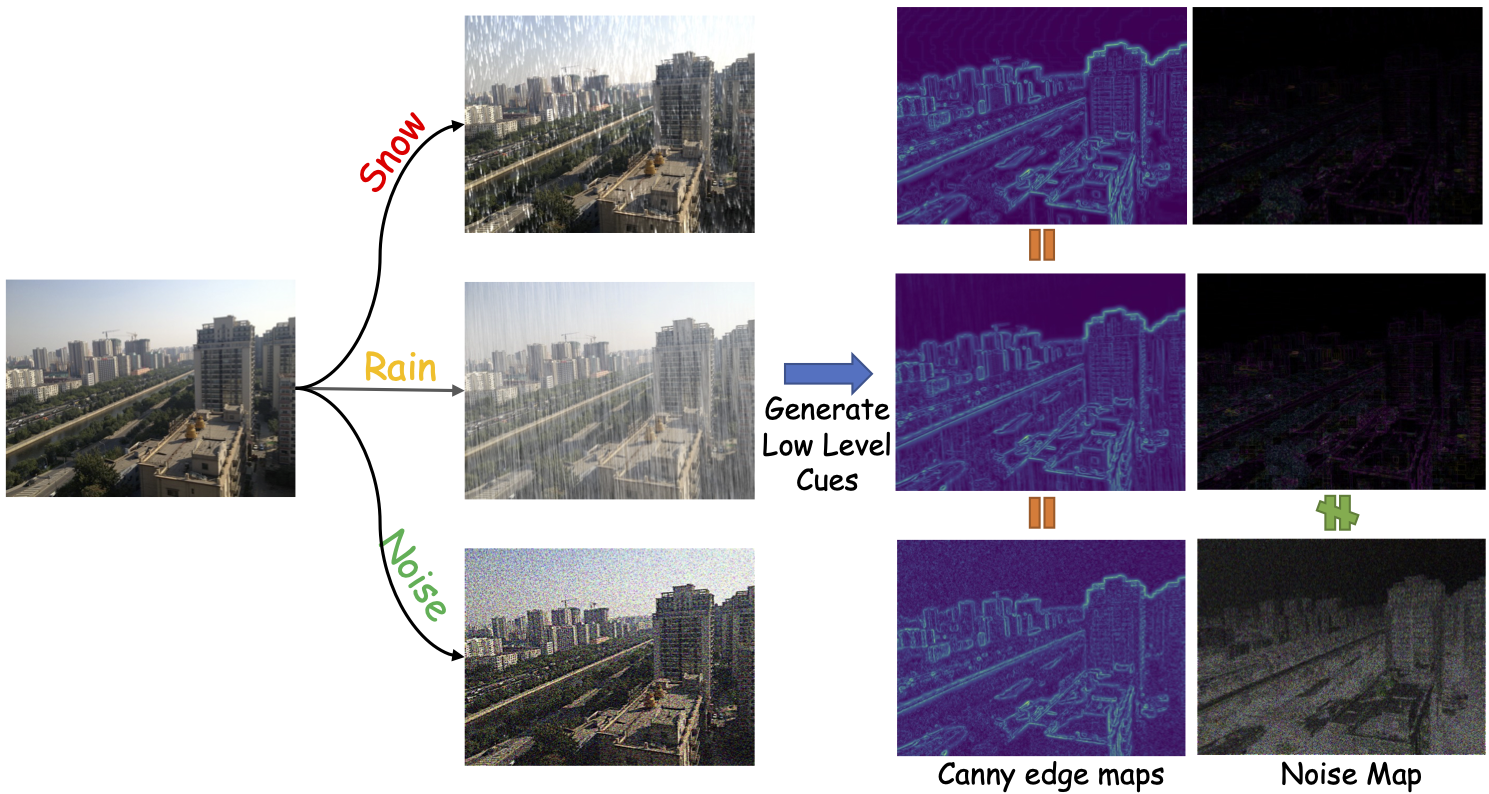} 
        \caption{Low level guidance}
        \label{fig:l_sub1}
    \end{subfigure}
    \hfill 
    \begin{subfigure}[b]{0.48\linewidth}
        \centering
        \includegraphics[width=\linewidth]{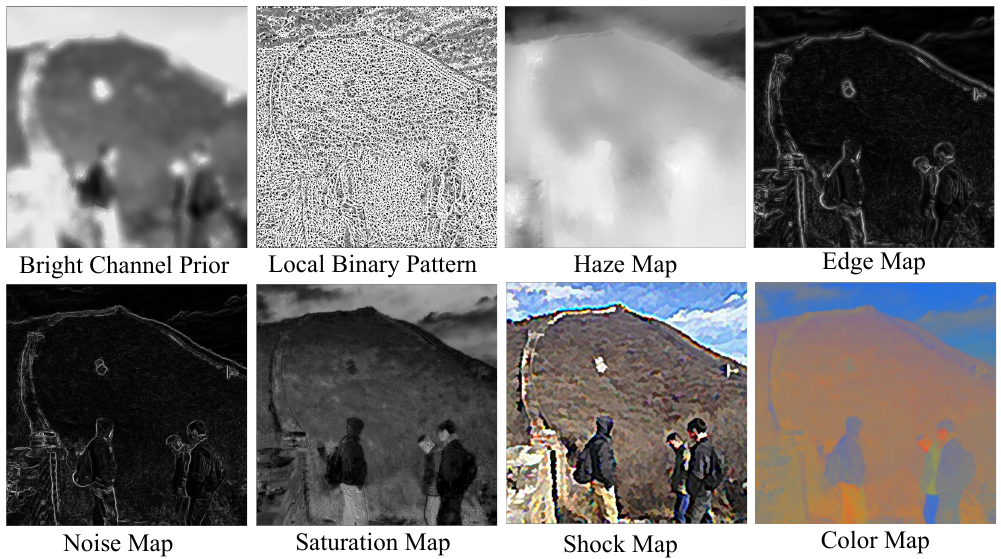} 
        \caption{All low level cues generated.}
        \label{fig:l_sub2}
    \end{subfigure}
    \caption{\textit{Low-level cues.} (a) Low-level cues provide discriminative signals across degradation types, enabling automatic restoration. (b) Our framework leverages diverse cues such as edges, color statistics, noise, and transmission, to characterize the input scene.}
    \label{fig:lowlev}
\end{figure}


\vspace{-0.25cm}
\subsection{\textsc{UnSCAR}}
UnSCAR is designed to scale universal restoration along two axes: \emph{representation scaling} and \emph{conditional capacity scaling}. 
At the representation level, we enhance degradation observability and discriminability while removing manual task-specific routing. 
At the architectural level, we redesign the conditional control pathway to scale to diverse degradations without proportional growth in parameters or computation. 
Finally, we introduce lightweight \emph{practical extensions} to improve restoration fidelity, controllability, and adaptability to new tasks.

\paragraph{\textbf{1) Unified and Degradation-Aware Guidance Representation.}}
As degradation diversity increases, reliable distinction among heterogeneous and mixed degradations becomes critical. 
As discussed in Sec.~\ref{sec:pre}, manual task-to-cue specification limits scalability when the degradation space expands. 
To resolve this, we replace task-specific cue pathways with a unified guidance encoder.

\paragraph{Richer Low-Level Structural Priors.}
To improve degradation observability, as shown in Fig.~\ref{fig:lowlev}, we expand the pool of low-level visual cues extracted from the degraded image $\tilde{I}$, approximately doubling the number used in UniCoRN.
These cues provide complementary structural signals across degradation types (e.g., noise amplification, haze patterns, color shifts), enhancing the model’s ability to identify degradation characteristics.

\begin{figure}[t]
    \centering
    \includegraphics[width=1\linewidth]{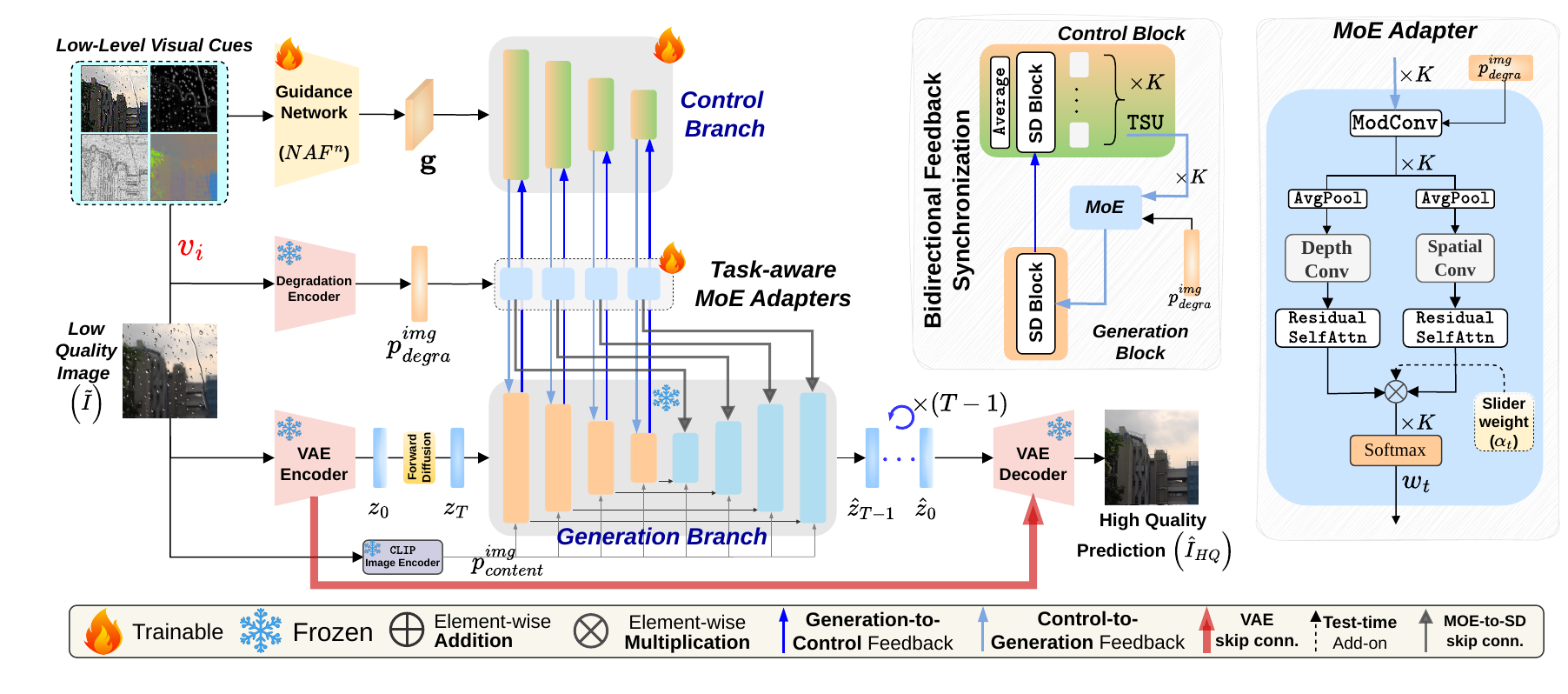}
    \caption{\textit{\modelName architecture.} We leverage low-level visual guidance within a bidirectional feedback, controllable generative network with mixture-of-experts. This enables unified image restoration across multiple simultaneous degradations.}
    \label{fig:mainfig}
\end{figure}

\paragraph{Unified Guidance Encoding.}
Instead of maintaining task-specific cue pathways, we adopt a unified guidance encoder that jointly processes the degraded image and all extracted cues. 
Each input (the degraded image $\tilde{I}$ and the enriched low-level cues $\{v_i\}$) is encoded independently through a lightweight chain of $n$ Nonlinear Activation Free $(NAF)$ layers~\cite{chen2022nafnet}. 
The resulting features are aggregated via summation before being passed to the diffusion backbone (see \cref{fig:mainfig}):
\vspace{-0.1cm}
\begin{equation}
    \mathbf{g} = {NAF}_0^n(\tilde{I}) + \sum_i {NAF}_i^n(v_i)\,,
\end{equation}
where $\mathbf{g}$ denotes the unified guidance representation shared across all degradation tasks. 
Here, $v_i$ represents the enriched low-level cues, and $i$ indexes the corresponding $NAF$ encoding branches. 
This unified encoding eliminates manual cue-to-task specification and enables the model to implicitly infer degradation composition from shared structural representations.

\begin{figure}[t]
    \centering
    \includegraphics[width=\linewidth]{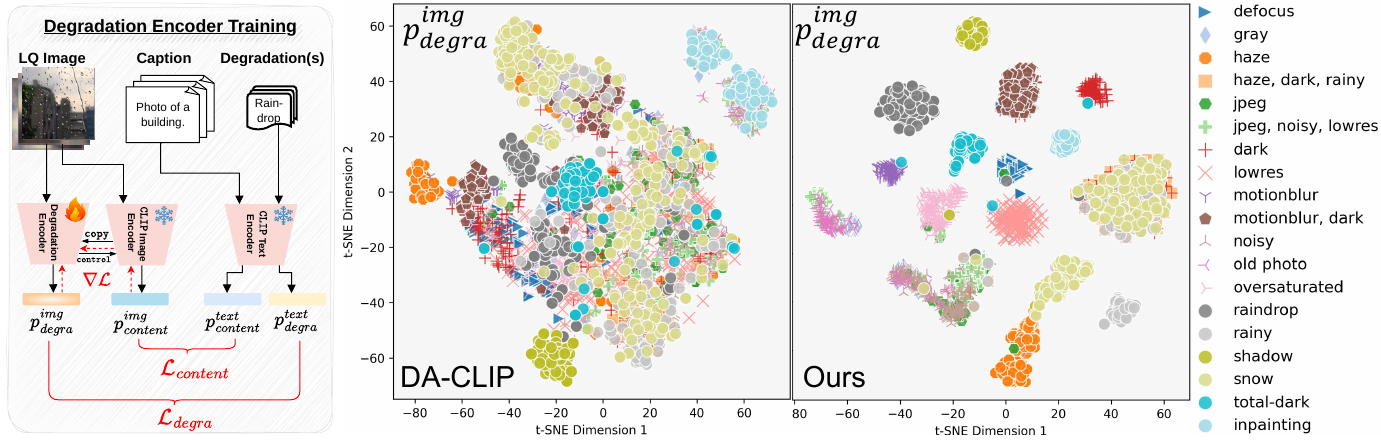}
    \caption{\textit{Degradation-aware embeddings.} Our degradation-aware training framework (left) learns an auxilliary degradation encoder cloned from the CLIP image encoder trained with SigLIP loss \cite{zhai2023sigmoid}. The resulting embeddings provide clear separation between single- and mixed-degradation tasks while maintaining proximity across related degradations.
    }
    \label{fig:embedding}
    \vspace{-0.1cm}
\end{figure}

\paragraph{Degradation-Aware Embedding Augmentation.}
As degradation diversity increases, precise task differentiation within the conditional pathway becomes essential to prevent feature interference. 
Control signals from heterogeneous tasks may overlap (e.g., rain+fog vs. rain or fog), resulting in ambiguous conditional modulation and degraded restoration quality, particularly under mixed-degradation settings.
To enhance task discriminability, we introduce degradation-aware embeddings built upon a visual-language model~\cite{luo2023controlling}. 
Specifically, we augment the CLIP image encoder~\cite{radford2021learning} with an additional degradation controller head, which disentangles content embeddings $p^{img}_{content}$ from degradation embeddings $p^{img}_{degra}$ (see \cref{fig:embedding}). 
Together, these designs establish a unified and degradation-aware guidance representation that scales to diverse and mixed-degradations.

\paragraph{\textbf{2) Capacity-Efficient ControlNet.}} 
To realize capacity-efficient conditioning, we enhance the conditional pathway through control signal compression, efficient mixture-of-experts adapters and  bidirectional feedback synchronization (\cref{fig:mainfig}).

\paragraph{Shared Control Compression.}
As the number of degradation tasks increases, naively forwarding each task-specific control signal $\mathcal{F}_t$ ($t$ indexes the degradation tasks) independently through heavy ControlNet layers $(L)$ results in redundant computation and quickly becomes a scalability bottleneck.
To maintain capacity-efficient conditioning, we introduce a lightweight \emph{squeeze-and-excite} mechanism that amortizes control computation across tasks. The \textit{squeeze} operation aggregates task-specific control features $\mathcal{F}_t$ via averaging and
producing a shared control representation $s^k$. The \textit{excitation} stage then redistributes this shared representation to each task through task-specific modulation units (TSU) \cite{mandal2025unicorn},
\begin{equation}
s^k = \texttt{Average}(\mathcal{F}_t^{k-1}) \,, \,\, \mathcal{F}_t^k = \texttt{TSU}_t(L^k(s^k))\,; \mathcal{F}^0_t = \textbf{g}\,.
\end{equation}
$L^k$ denotes $k$-th shared control layer.
This design maintains task-specific modulation while avoiding repeated control computations, enabling improved effective conditional capacity under limited parameter capacity and compute budgets. 


\paragraph{Scaling the Mixture-of-Experts Adapter.}
Many prior all-in-one restoration methods rely on inference-time weighting to balance restoration trajectories~\cite{mandal2025unicorn, unires}. In contrast, we integrate task balancing directly within the mixture-of-experts (MoE) routing mechanism. 
While UniCoRN employs multi-dimensional spatially-varying routing, such routing becomes increasingly complex and difficult to interpret as the number of degradation types grows. 
We instead adopt a global task-level routing strategy that predicts a single normalized weight vector over degradation tasks.
Specifically, we compress $\mathcal{F}_t$ via spatial average pooling with factor $\downarrow f$, yielding a downsampled representation of size $(H\times W)\to (\frac{H}{f}\times \frac{W}{f})$, which is subsequently processed through depth-wise and spatial convolutions~\cite{mandal2025unicorn}, followed by residual attention blocks~\cite{xinsir2024controlnetplus}. The resulting features are projected to a task-level weight vector $\mathbf{w} \in \mathbb{R}^T$, normalized via softmax to enforce $\sum_t w_t^k = 1$.
To further enhance degradation awareness, each expert router $E^k$ is conditioned on the degradation embeddings $p^{img}_{degra}$ via style modulation~\cite{karras2020analyzing}:
\begin{equation}
    \mathcal{F}^k_\text{fused}
    =
    \sum
    \underbrace{E^k((\mathcal{F}_t^k)\downarrow_f; p^{img}_{degra})}_{w^k_t}
    \cdot
    \mathcal{F}_t^k.
\end{equation}
$\mathcal{F}_\text{fused}^k$ denotes the fused feature at layer $k$.
This global routing improves conditional capacity while maintaining parameter and computational efficiency, enabling scalable multi-degradation restoration.

\paragraph{Bidirectional Feedback Synchronization.}
UniCoRN follows the original ControlNet paradigm, where feedback signals are injected only into the decoder layers of the diffusion backbone. The backbone encoder remains frozen and processes degraded inputs without intermediate correction.
Under this decoder-only injection design, degradation-induced artifacts accumulate in encoder representations before any corrective modulation is applied. The decoder-side control branch must then compensate for these errors retroactively, increasing correction strain and limiting effective conditional capacity as degradation diversity grows.
To alleviate this structural imbalance, we adopt a bidirectional synchronization mechanism, inspired by \cite{zavadski2024controlnet}, that enables feature exchange between the control encoder to the frozen backbone encoder:
\begin{equation}
    \mathcal{F}^{k+1}_{fused} = L_{c}^{k+1}(\mathcal{F}_\text{fused}^k + \mathcal{F}_{b}^k), \quad
    \mathcal{F}^{k+1}_b = L_b^{k+1}(\mathcal{F}_\text{fused}^k + \mathcal{F}_b^k),
\end{equation}
where $\mathcal{F}_b^k$ and $\mathcal{F}_c^k$ denote backbone and control signal at layer $k$.
This early feature exchange distributes correction across the network, increasing effective conditional capacity yet keeping a lightweight control branch ($\sim 0.2\times$ less parameters).


\paragraph{3) Practical Extensions.}
Complementing the representation and control redesign that address scalability, we further introduce lightweight extensions to enhance fidelity and expand the practical usability of the framework.


\paragraph{\textbf{Residual VAE Refinement.}}
Due to the lossy nature of the variational autoencoder (VAE) used in SD 1.5 \cite{rombach2022high}, we introduce a simple modification to the decoder architecture to better suit image restoration, where faithful reconstruction is required rather than hallucination.
Specifically, we forward the intermediate activations of the degraded input image from the VAE encoder to the decoder after the diffusion-based correction is performed in our model. 
These encoder features are then alpha-blended with the decoder’s upsampled activations at each level using zero-initialized residual convolutional blocks.

\paragraph{\textbf{Controllable Restoration via Sliders.}}
Controlling the image restoration process enhances model customizability for photo editing applications and provides greater transparency into its working principles, in contrast to prior black-box all-in-one restoration methods.
We introduce a controllability mechanism within our universal restoration framework by associating each restoration task (e.g., deblurring, denoising, dehazing) with an independent user-controlled intensity slider. 
This design enables fine-grained control over different degradation types rather than relying on a single global restoration strength.
Formally, let $w_t$ denote the fusion control weight generated by the mixture-of-expert adapter for a specific degradation type ($t \in T$). For each degradation, the user also specifies an intensity coefficient $\alpha_t \in [0,1]$.
The controlled fusion weights at layer $k$ are computed as
$
\tilde{w}^k_t = \frac{\alpha_t w^k_t}{\sum_{T} \alpha_t w^k_t},
$
which ensures $\tilde{w}^k_t \in [0,1]$ and $\sum_t \tilde{w}^k_t = 1$.
This task-wise re-weighting preserves the scale and stability of the latent representation while allowing continuous and independent control over each degradation.

\paragraph{Few-shot unseen task Adaptation.}
While our model generalizes to unseen degradations in natural images, adapting to significantly different Out-of-domain (OOD) images (e.g., medical imaging) remains challenging.
Leveraging the modular design of our architecture, we enable efficient OOD adaptation by introducing an auxiliary low-level encoder head (NAF chain) that processes degraded images alongside the unified low-level encoder described before. 
For each new OOD dataset, we also duplicate a task-specific unit (TSU) similar to few-shot task adaptation case and its associated squeeze-excitation modules at each encoder layer, leaving the core mixture-of-experts blocks untouched. 
This lightweight extension enables effective adaptation to challenging domains, such as medical imaging, without disrupting previously learned restoration capabilities or inducing catastrophic forgetting.

\vspace{0.05cm}

    

\vspace{-0.2cm}
\begin{figure}[h!]
    \centering
    \includegraphics[width=\linewidth]{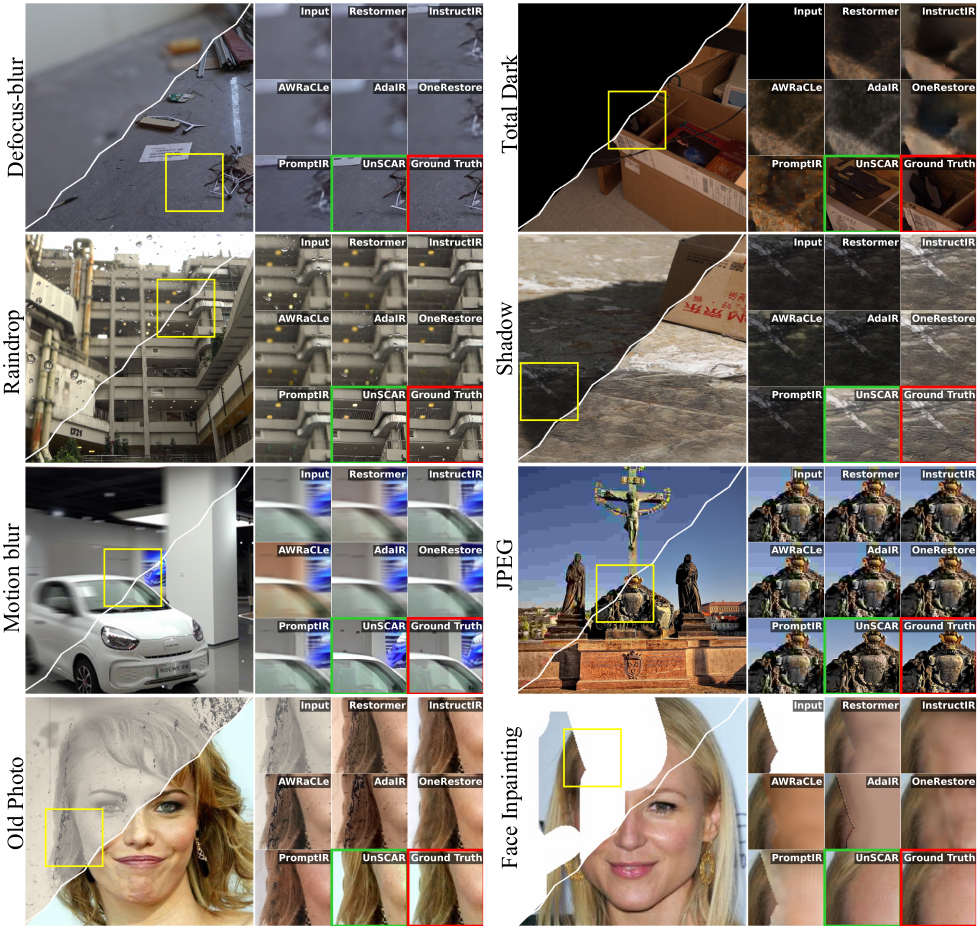}
    \caption{\textit{Single-degradation restoration.} \modelName produces sharper, higher-fidelity reconstructions across single-degradation tasks.}
    \label{fig:single_degra}
\end{figure}
\subsection{Training Objectives and Strategy}
\label{sec:optim}
\vspace{-0.1cm}
During training, we adopt a training objective that enforces reconstruction fidelity and condition embedding supervision, and a routing stabilization strategy that stabilizes expert specialization.

\paragraph{\textbf{Training Objectives.}}
The overall objective is defined as,
$
    \mathcal{L}_{\text{total}} = \mathcal{L}_{\text{LDM}} + \lambda_{\text{pixel}} \mathcal{L}_{\text{pixel}},
$
where $\mathcal{L}_{\text{LDM}}$ is the latent diffusion training objective, following the DDIM formulation~\cite{song2020denoising} with $\epsilon$-parameterization:
\begin{equation}
    \mathcal{L}_{\text{diff}}
    =
    \mathbb{E}_{t, \epsilon}
    \left[
    \| \epsilon - \epsilon_\theta(x_t, t) \|^2
    ;\{\mathcal{F}_\text{fused}^k\}\right],
\end{equation}
where $\epsilon \sim \mathcal{N}(0,1)$ denotes Gaussian noise, $x_t$ is the noisy latent at timestep $t$, $\{\mathcal{F}_\text{fused}^k\}$ is the set of control feature from all intermediate layers and $\epsilon_\theta(x_t,t)$ predicts the noise. $\mathcal{L}_\text{pixel}$ is a \emph{Pixel-Space Reconstruction Loss}, which is introduced to mitigate hallucinated textures and structural artifacts. $\mathcal{L}_\text{pixel}$ is defined as:
\begin{equation}
    \mathcal{L}_{\text{pixel}} = \gamma_1\| x_\text{gt} - \mathcal{D_\phi}(\hat{z}_0) \|_1 + \gamma_2\mathcal{L}_\texttt{SSIM} (x_\text{gt}, \mathcal{D_\phi}(\hat{z}_0)) + \gamma_3\mathcal{L}_\texttt{LPIPS} (x_\text{gt}, \mathcal{D_\phi}(\hat{z}_0)),
\end{equation}
where $x_\text{gt}$ is the ground-truth image, $\mathcal{D_\phi}$ denotes the VAE decoder with encoder skip-connections, and $\hat{z}_0$ is the predicted clean latent from noise ($\epsilon(x_t, t)$) at time step t. $\lambda_{\text{pixel}}$ and $\gamma_{1,2,3}$ are a balancing weights. $\lambda_{\text{pixel}}=\gamma_1=\gamma_3=0.4$, $\gamma_2=0.2$.

\paragraph{Condition Embedding Supervision.}
To enable accurate degradation discrimination and task-aware routing, we supervise condition embeddings via
\begin{equation}
    \mathcal{L}_{\text{cond}} = \mathcal{L}_{\text{content}}(p^{img}_{content})  + \mathcal{L}_{\text{degradation}}(p^{img}_{degra}).
\end{equation}
$p^{img}_{content}$ and $p^{img}_{degra}$ denote content and degradation embeddings respectively. 
Prior works such as DA-CLIP~\cite{luo2023controlling} rely on symmetric InfoNCE objectives \cite{radford2021learning} for both content and degradation losses to enforce strict one-to-one alignment between images and degradation captions. However, mixed degradations often share similarities across multiple categories, making hard contrastive pairing suboptimal. 
We adopt the \textbf{sigmoid loss} from SigLIP~\cite{zhai2023sigmoid} for $\mathcal{L}_{\text{degradation}}$, which independently scores each image-caption pair via sigmoid activation (see \cref{fig:embedding}).

\paragraph{\textbf{Expert Routers Stabilization Strategy.}}
To prevent expert routers from converging to suboptimal local minima or collapsing into a premature ``winner-takes-all'' regime, we employ a two-stage curriculum training strategy with label smoothing. Instead of using hard routing targets, we introduce controlled entropy into the routing distribution.
During the initial exploration phase, we assign $\alpha \approx 0.8$ probability mass uniformly across task-relevant experts and distribute the remaining $(1-\alpha)$ mass among suboptimal experts. This higher-entropy target maintains gradient flow across routing paths, mitigating dead-expert collapse and encouraging robust specialization.
In the fine-tuning phase, the target mass for optimal experts is increased to $0.95$, reducing entropy and promoting sparse, confident routing decisions. 
This curriculum progressively transitions the router from exploratory soft assignment to stable, task-discriminative expert selection.
\textit{Additional implementation details are provided in the supplementary material.}

\vspace{-2mm}
\section{Experiments and Results}
\label{sec:experiments}
\vspace{-2mm}

\begin{figure}[t]
    \centering
    \includegraphics[width=\linewidth]{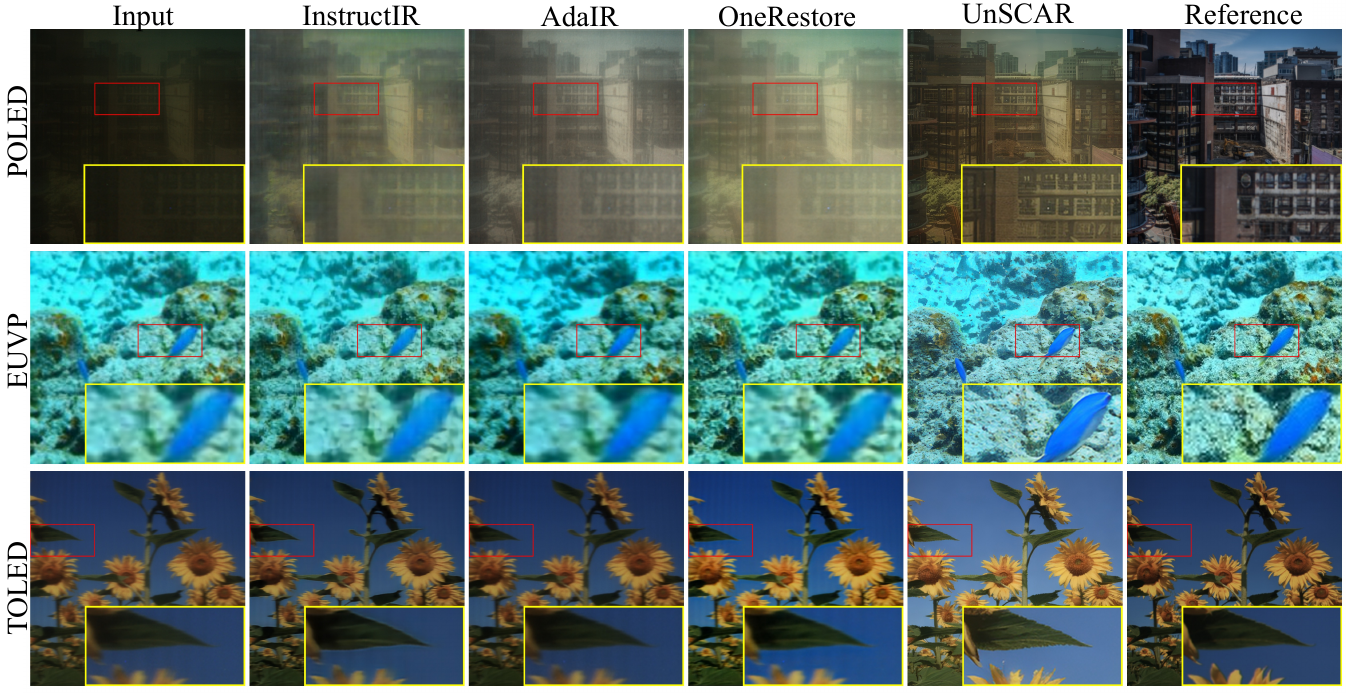}
    \caption{\textit{Generalization to unseen degradations.} Zero-shot evaluation of baseline methods and our approach on unseen mixed-degradation datasets.}
    \label{fig:unseen}
\end{figure}
\subsection{Dataset and Metrics}
\label{sec:dataset} 
We collect and curate a large-scale paired degradation dataset comprising 16 single-degradation, 10 mixed-degradation, and 3 real-world degradation (used only for evaluation) datasets. Detailed dataset composition is provided in the supplementary.
Training data predominantly consist of natural outdoor images exhibiting real-world degradations, paired with clean targets.
To evaluate performance under mixed degradations, we additionally curate synthetic combinations including (low light, rain, haze) from CDD~\cite{guo2024onerestore}, (low light, blur, noise) from LOL-Blur~\cite{zhou2022lednet}, and (low resolution, low light) from RELLISUR~\cite{aakerberg2021rellisur}. 
All methods are evaluated using pixel-wise (i.e. PSNR, SSIM) and perceptual metrics (LPIPS~\cite{lpips}, DISTS~\cite{ding2020image}). 
We further report non-reference IQA metrics (NIQE~\cite{niqe}, BRISQUE~\cite{brisque}) to evaluate on completely unseen degradations.


\begin{table*}[tp]
  \caption{\textit{Single-degradation restoration performance.} All models are trained jointly on all tasks to evaluate scalability. \textbf{Bold} denotes best performance; \underline{underline} indicates second best. ${\textcolor{blue}{\blacksquare}}$ and  ${\textcolor{olive}{\blacksquare}}$ denotes prior single-task and all-in-one restoration models.
  }
  \vspace{-0.2cm}
  \label{tab:air_degrade}
  \centering
  \fontsize{8pt}{\baselineskip}\selectfont 
  \setlength\tabcolsep{3pt} 
  \resizebox{\textwidth}{!}
  {
  \begin{tabular}{ c c|ccc|ccc|ccc|ccc }
    \hline 

    \multicolumn{2}{c|}{} & \multicolumn{3}{c|}{\textbf{\textit{Dehazing:}} RESIDE \cite{reside}} & \multicolumn{3}{c|}{\textbf{\textit{Motion Deblur}}: RHM \cite{zhang2023benchmarking}} & \multicolumn{3}{c|}{\textbf{\textit{Low-Light:}} LOL \cite{wei2018deep}} & \multicolumn{3}{c}{\textbf{\textit{Denoising}}: CBSD68 \cite{cbsd68}} \\ 
    \cline{3-14}
    \multicolumn{1}{c}{} & \textbf{Method} & PSNR $\uparrow$ & SSIM $\uparrow$ & LPIPS $\downarrow$ & PSNR $\uparrow$ & SSIM $\uparrow$ & LPIPS $\downarrow$ & PSNR $\uparrow$ & SSIM $\uparrow$ & LPIPS $\downarrow$ & PSNR $\uparrow$ & SSIM $\uparrow$ & LPIPS $\downarrow$ \\ 
    \hline

    \multirow{2}{*}{\large$\textcolor{blue}{\blacksquare}$} 
    & Restormer \cite{zamir2022restormer} & 26.57 & 0.95 & 0.05 & 26.41 & 0.81 & 0.35 & 20.43 & 0.83 & 0.30 & 28.60 & 0.79 & 0.23 \\
    & NAFNet \cite{chen2022nafnet} & 27.62 & 0.96 & 0.04 & 27.87 & 0.84 & 0.24 & 22.19 & 0.88 & 0.16 & \textbf{32.71} & \textbf{0.89} & \textbf{0.08} \\
    \cline{2-14} 
    \multirow{5}{*}{\large$\textcolor{olive}{\blacksquare}$} 
    & PromptIR \cite{potlapalli2024promptir} & 26.27 & 0.95 & \underline{0.03} & \underline{28.69} & \underline{0.86} & \underline{0.18} & 18.58 & 0.83 & 0.20 & 30.52 & 0.84 & 0.16 \\
    & OneRestore \cite{guo2024onerestore} & \underline{30.68} & \underline{0.97} & \textbf{0.02} & 27.95 & 0.82 & 0.27 & 22.21 & 0.88 & 0.19 & \underline{32.27} & \underline{0.88} & 0.11 \\
    & InstructIR \cite{instructir} & 29.78 & \underline{0.97} & \textbf{0.02} & 27.14 & 0.83 & 0.24 & \underline{22.33} & \underline{0.89} & \underline{0.14} & 32.10 & \underline{0.88} & 0.10 \\
    & AdaIR \cite{adair} & 24.71 & 0.92 & 0.07 & 26.34 & 0.79 & 0.30 & 17.26 & 0.82 & 0.33 & 25.43 & 0.76 & 0.27 \\
    & AwRaCle \cite{rajagopalan2024awracle} & 25.65 & 0.93 & 0.07 & 25.27 & 0.80 & 0.26 & 17.95 & 0.80 & 0.31 & 26.79 & 0.77 & 0.27 \\
    \hline
    \rowcolor{YellowGreen!40}
    & \textbf{\textsc{UnSCAR}} & \textbf{31.89} & \textbf{0.98} & \underline{0.03} & \textbf{29.44} & \textbf{0.92} & \textbf{0.16} & \textbf{22.51} & \textbf{0.93} & \textbf{0.12} & 31.85 & \underline{0.88} & \underline{0.09} \\
    
    \hline 

    \multicolumn{2}{c|}{} & \multicolumn{3}{c|}{\textbf{\textit{Defocus Deblur:}} LSD \cite{zhang2023benchmarking}} & \multicolumn{3}{c|}{\textbf{\textit{Desnowing}}: SRRS \cite{chen2023snow}} & \multicolumn{3}{c|}{\textbf{\textit{Deraining}}: Rain4K \cite{chen2026towards}} & \multicolumn{3}{c}{\textbf{\textit{DeRaindrop}}: Raindrop \cite{jin2024raindrop}} \\ 
    \cline{3-14}
    \multicolumn{1}{c}{} & \textbf{Method} & PSNR $\uparrow$ & SSIM $\uparrow$ & LPIPS $\downarrow$ & PSNR $\uparrow$ & SSIM $\uparrow$ & LPIPS $\downarrow$ & PSNR $\uparrow$ & SSIM $\uparrow$ & LPIPS $\downarrow$ & PSNR $\uparrow$ & SSIM $\uparrow$ & LPIPS $\downarrow$ \\ 
    \hline

    \multirow{2}{*}{\large$\textcolor{blue}{\blacksquare}$} 
    & Restormer \cite{zamir2022restormer} & 22.35 & 0.80 & 0.56 & 28.52 & 0.92 & 0.09 & 26.85 & 0.84 & 0.23 & 20.57 & 0.58 & 0.53 \\
    & NAFNet \cite{chen2022nafnet} & 22.61 & \underline{0.83} & \underline{0.35} & 28.84 & 0.93 & 0.07 & \underline{28.38} & \underline{0.89} & 0.16 & \underline{23.75} & \underline{0.71} & \underline{0.35} \\
    \cline{2-14} 
    \multirow{5}{*}{\large$\textcolor{olive}{\blacksquare}$} 
    & PromptIR \cite{potlapalli2024promptir} & 22.30 & 0.74 & 0.40 & 29.15 & \underline{0.94} & \underline{0.06} & \textbf{29.80} & \textbf{0.92} & \textbf{0.08} & 21.97 & 0.62 & 0.44 \\
    & OneRestore \cite{guo2024onerestore} & 22.75 & 0.75 & 0.43 & 28.98 & 0.93 & 0.08 & 26.51 & 0.84 & 0.16 & 22.43 & 0.65 & 0.45 \\
    & InstructIR \cite{instructir} & \underline{23.52} & 0.76 & 0.39 & 29.04 & \underline{0.94} & 0.07 & 25.22 & 0.84 & 0.16 & 22.33 & 0.65 & 0.46 \\
    & AdaIR \cite{adair} & 20.09 & 0.68 & 0.56 & 28.44 & 0.92 & 0.08 & 23.85 & 0.74 & 0.26 & 18.87 & 0.50 & 0.53 \\
    & AwRaCle \cite{rajagopalan2024awracle} & 20.60 & 0.70 & 0.55 & \underline{29.22} & \underline{0.94} & \underline{0.06} & 25.06 & 0.78 & 0.23 & 19.07 & 0.53 & 0.56 \\
    \hline
    \rowcolor{YellowGreen!40}
    & \textbf{\textsc{UnSCAR}} & \textbf{23.62} & \textbf{0.84} & \textbf{0.25} & \textbf{31.39} & \textbf{0.97} & \textbf{0.03} & \underline{29.30} & \underline{0.91} & \underline{0.09} & \textbf{24.03} & \textbf{0.72} & \textbf{0.31} \\
    
    \hline 

    \multicolumn{2}{c|}{} & \multicolumn{3}{c|}{\textbf{\textit{Overexposure:}} EED \cite{afifi2021learning}} & \multicolumn{3}{c|}{\textbf{\textit{Total Dark}}: Sony TDD \cite{yan2024you}} & \multicolumn{3}{c|}{\textbf{\textit{JPEG:}} Flickr2K \cite{Agustsson_2017_CVPR_Workshops}} & \multicolumn{3}{c}{\textbf{\textit{Superresolution:}} RealSR \cite{cai2019toward}} \\ 
    \cline{3-14}
    \multicolumn{1}{c}{} & \textbf{Method} & PSNR $\uparrow$ & SSIM $\uparrow$ & LPIPS $\downarrow$ & PSNR $\uparrow$ & SSIM $\uparrow$ & LPIPS $\downarrow$ & PSNR $\uparrow$ & SSIM $\uparrow$ & LPIPS $\downarrow$ & PSNR $\uparrow$ & SSIM $\uparrow$ & LPIPS $\downarrow$ \\ 
    \hline

    \multirow{2}{*}{\large$\textcolor{blue}{\blacksquare}$} 
    & Restormer \cite{zamir2022restormer} & 20.75 & 0.88 & 0.15 & 19.02 & 0.64 & 0.62 & 27.43 & 0.88 & 0.12 & 32.77 & 0.92 & 0.13 \\
    & NAFNet \cite{chen2022nafnet} & 21.32 & 0.91 & \underline{0.10} & 20.88 & 0.69 & \underline{0.53} & 28.93 & 0.89 & 0.13 & \textbf{38.70} & \textbf{0.96} & \textbf{0.06} \\
    \cline{2-14} 
    \multirow{5}{*}{\large$\textcolor{olive}{\blacksquare}$} 
    & PromptIR \cite{potlapalli2024promptir} & 17.62 & 0.87 & 0.15 & 20.43 & 0.62 & 0.58 & 29.34 & 0.90 & 0.12 & 36.01 & \underline{0.94} & \underline{0.09} \\
    & OneRestore \cite{guo2024onerestore} & 20.90 & \underline{0.92} & \underline{0.10} & \underline{21.46} & \underline{0.71} & 0.57 & 29.61 & 0.90 & 0.12 & 34.11 & 0.93 & \underline{0.09} \\
    & InstructIR \cite{instructir} & \underline{21.57} & \underline{0.92} & \underline{0.10} & 20.29 & 0.69 & 0.55 & \underline{29.65} & \underline{0.91} & \underline{0.10} & 36.73 & 0.93 & \underline{0.09} \\
    & AdaIR \cite{adair} & 19.77 & 0.90 & 0.14 & 18.84 & 0.61 & 0.65 & 27.09 & 0.88 & 0.13 & 31.98 & 0.92 & 0.13 \\
    & AwRaCle \cite{rajagopalan2024awracle} & 17.65 & 0.86 & 0.17 & 19.47 & 0.55 & 0.64 & 27.68 & 0.88 & 0.14 & 31.35 & 0.92 & 0.14 \\
    \hline
    \rowcolor{YellowGreen!40}
    & \textbf{\textsc{UnSCAR}} & \textbf{22.55} & \textbf{0.96} & \textbf{0.08} & \textbf{21.92} & \textbf{0.74} & \textbf{0.37} & \textbf{30.84} & \textbf{0.95} & \textbf{0.07} & \underline{37.36} & \underline{0.94} & \underline{0.07} \\

    \hline \hline

    \multicolumn{2}{c|}{\textbf{\textit{(Generative Tasks)}}} & \multicolumn{3}{c}{\cellcolor{yellow!20}\textbf{\textit{Inpainting}}: CelebA \cite{liu2015faceattributes}} & \multicolumn{3}{c}{\cellcolor{yellow!20}\textbf{\textit{Old Photo:}} MLRN \cite{MLRN}} & \multicolumn{3}{c}{\cellcolor{yellow!20}\textbf{\textit{Gray2Color}}: Div2K \cite{div2k}} & \multicolumn{3}{c}{\cellcolor{yellow!20}\textbf{\textit{Deshadowing}} SRD \cite{guo2024single}} \\ 
    \cline{3-14}
    \multicolumn{2}{c|}{\textbf{Method}} & PSNR $\uparrow$ & SSIM $\uparrow$ & LPIPS $\downarrow$ & PSNR $\uparrow$ & SSIM $\uparrow$ & LPIPS $\downarrow$ & PSNR $\uparrow$ & SSIM $\uparrow$ & LPIPS $\downarrow$ & PSNR $\uparrow$ & SSIM $\uparrow$ & LPIPS $\downarrow$ \\ 
    \hline

    \multirow{2}{*}{\large$\textcolor{blue}{\blacksquare}$} 
    & Restormer \cite{zamir2022restormer} & 22.03 & 0.85 & 0.28 & 18.37 & 0.84 & 0.35 & 21.70 & 0.92 & 0.21 & 18.71 & 0.82 & 0.18 \\
    & NAFNet \cite{chen2022nafnet} & 26.64 & 0.89 & 0.15 & 27.08 & 0.92 & 0.16 & 27.97 & 0.93 & 0.13 & 28.96 & \underline{0.90} & \underline{0.09} \\
    \cline{2-14} 
    \multirow{5}{*}{\large$\textcolor{olive}{\blacksquare}$} 
    & PromptIR \cite{potlapalli2024promptir} & 23.25 & 0.88 & 0.21 & 21.35 & 0.89 & 0.22 & 25.80 & 0.94 & 0.14 & 22.51 & 0.86 & 0.14 \\
    & OneRestore \cite{guo2024onerestore} & 27.01 & \underline{0.91} & 0.16 & 27.11 & 0.94 & 0.17 & 27.63 & 0.94 & 0.14 & \underline{29.64} & 0.89 & 0.10 \\
    & InstructIR \cite{instructir} & \underline{27.12} & \underline{0.91} & \underline{0.15} & \textbf{29.06} & \textbf{0.96} & \textbf{0.11} & \textbf{29.68} & \textbf{0.98} & \textbf{0.07} & 29.61 & \underline{0.90} & 0.10 \\
    & AdaIR \cite{adair} & 18.04 & 0.78 & 0.39 & 17.81 & 0.83 & 0.36 & 22.05 & 0.85 & 0.21 & 16.98 & 0.80 & 0.19 \\
    & AwRaCle \cite{rajagopalan2024awracle} & 22.99 & 0.88 & 0.24 & 19.80 & 0.84 & 0.32 & 24.51 & 0.89 & 0.16 & 21.30 & 0.85 & 0.17 \\
    \hline
    \rowcolor{YellowGreen!40}
    & \textbf{\textsc{UnSCAR}} & \textbf{27.78} & \textbf{0.97} & \textbf{0.08} & \underline{28.70} & \underline{0.93} & \underline{0.13} & \underline{28.15} & \underline{0.96} & \underline{0.09} & \textbf{31.32} & \textbf{0.95} & \textbf{0.05} \\
    \hline
  \end{tabular}
  }
\end{table*}

\vspace{-2mm}
\subsection{SOTA comparison on Single-Degradation Removal}
\vspace{-0.5mm}
\label{sec:comp_air} 
We first compare {\modelName} with prior universal restoration models \cite{guo2024onerestore, rajagopalan2024awracle, potlapalli2024promptir} on single-degradation restoration benchmarks \cite{reside, wei2018deep, cbsd68, chen2023snow, chen2026towards}. For fair comparison, all competing models are trained on the same unified training corpus as {\modelName}. Results in \cref{tab:air_degrade} show that {\modelName} achieves state-of-the-art performance on $11$ out of $16$ benchmarks across all metrics, while remaining highly competitive on the others. 
These results highlight the strong scalability of {\modelName} across diverse restoration tasks, where prior methods often become suboptimal when trained jointly on multiple degradations. 
Qualitative comparisons in \cref{fig:single_degra} further demonstrate that {\modelName} restores fine details more faithfully across tasks, including tiny objects (defocus and motion deblurring), distinct hair strands (face inpainting), and intricate building structures (raindrop removal). Overall, these results establish {\modelName} as a new state-of-the-art universal restoration model with strong generalization across diverse degradation types.

\vspace{-2mm}
\subsection{ Unseen, Mixed-degradation and In-the-Wild Restoration}
\label{sec:comp_mixed} 
\vspace{-1mm}


\begin{figure}[t]
    \centering
    \includegraphics[width=0.9\linewidth]{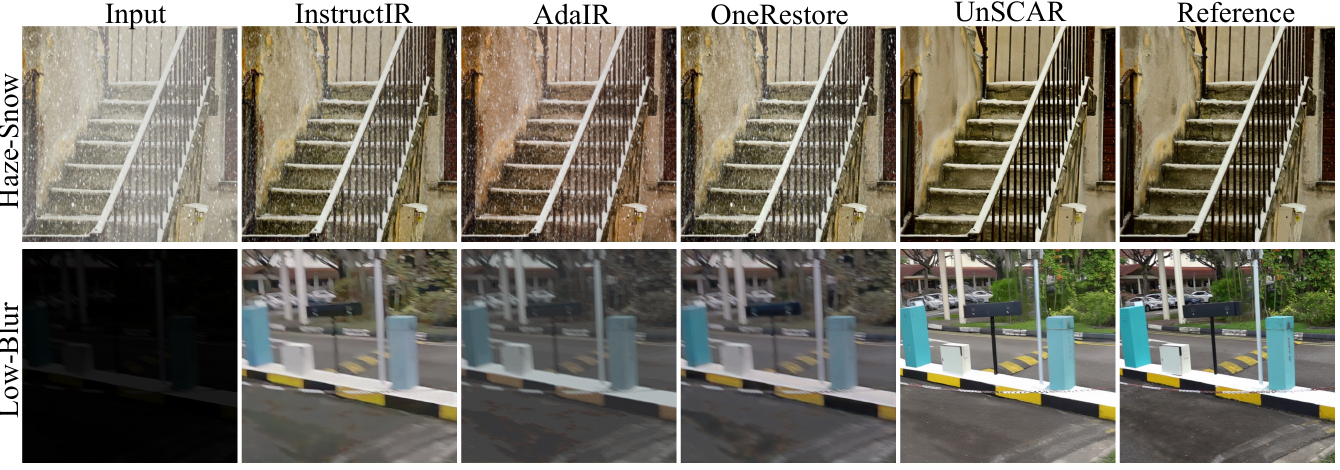}
    \caption{\textit{Mixed restoration.} Zero-shot evaluation of baseline methods vs ours for mixed restoration dataset with individual tasks seen during training.}
    \label{fig:mixed}
\end{figure}
\begin{figure}[t]
    \centering
    \includegraphics[width=.92\linewidth]{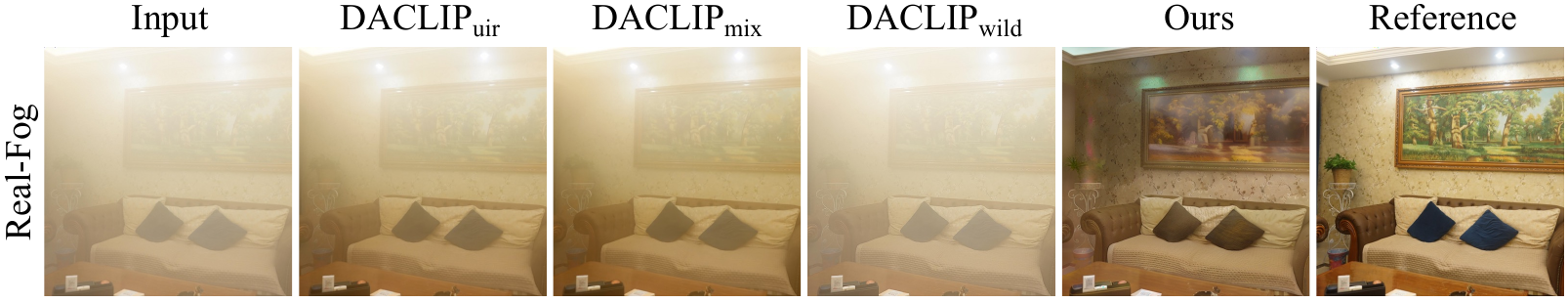}
    \caption{\textit{In-the-wild degradation removal.} Comparison with the diffusion based DA-CLIP method on heavy indoor smoke removal \cite{REVIDE}.}
    \label{fig:itw}
\end{figure}

\begin{table}[htbp]
    \centering
    \caption{Zero-shot performance on unseen-mixed (left) and mixed datasets (right). The best performing models across all tasks from Table 1 were used for evaluations. }
    \label{tab:unseen_mixed}
    \vspace{-0.3cm} 

    \begin{minipage}[t]{0.59\textwidth}
        \centering
        \resizebox{\textwidth}{!}{ 
            \begin{tabular}{l | cc | cc | cc}
                \hline
                \multirow{2}{*}{\textbf{Method}} & \multicolumn{2}{c|}{\textbf{TOLED \cite{zhou2021image}}} & \multicolumn{2}{c|}{\textbf{POLED \cite{zhou2021image}}} & \multicolumn{2}{c}{\textbf{EUVP \cite{islam2020fast}}} \\
                \cline{2-7} 
                & NIQE$\downarrow$ & BRISQUE$\downarrow$ & NIQE$\downarrow$ & BRISQUE$\downarrow$ & NIQE$\downarrow$ & BRISQUE$\downarrow$ \\
                \hline
                Restormer \cite{zamir2022restormer}  & 6.95 & 46.69 & 10.38 & 27.58 & 7.30 & 57.93 \\
                NAFNet \cite{chen2022nafnet}  & 6.90 & 49.58 & 6.65 & 34.42 & 6.76 & 50.37 \\
                OneRestore \cite{guo2024onerestore} & 6.93 & 52.97 & 6.49 & 16.57 & 7.43 & 56.45 \\
                InstructIR \cite{instructir} & 7.16 & 49.05 & 7.06 & 35.08 & 6.83 & 50.89 \\
                AdaIR \cite{adair} & 7.07 & 44.99 & 6.96 & 25.06 & 7.40 & 54.60 \\
                Prompt-IR \cite{potlapalli2024promptir} & 5.67 & 36.86 & 8.42 & 23.42 & 6.58 & 43.83 \\
                AwRaCle \cite{rajagopalan2024awracle} & 5.61 & 34.77 & 8.59 & \textbf{15.56} & 7.36 & 52.31 \\
                \rowcolor{YellowGreen!40}
                \textbf{UnSCAR} & \textbf{3.97} & \textbf{25.13} &  \textbf{5.21} & 16.21 & \textbf{3.58} & \textbf{20.77} \\
                \hline
            \end{tabular}
        }
    \end{minipage}
    \hfill 
    \begin{minipage}[t]{0.38\textwidth}
        \centering
        \resizebox{\textwidth}{!}{ 
           \begin{tabular}{l | cc | cc}
    \hline
    \multirow{2}{*}{\textbf{Method}} & \multicolumn{2}{c|}{\textbf{Haze+Snow} \cite{guo2024onerestore}} & \multicolumn{2}{c}{\textbf{Low-Blur} \cite{zhou2022lednet}} \\
    \cline{2-5} 
    & PSNR$\uparrow$ & LPIPS$\downarrow$ & PSNR$\uparrow$ & LPIPS$\downarrow$ \\
    \hline
    Restormer \cite{zamir2022restormer} & 17.68 & 0.222 & 13.92 & 0.381 \\
    NAFNet \cite{chen2022nafnet} & 17.40 & 0.186 & 16.45 & 0.312 \\
    OneRestore \cite{guo2024onerestore} & 17.00 & 0.210 & 12.84 & 0.402 \\
    InstructIR \cite{instructir} & 16.91 & 0.219 & 15.31 & 0.334 \\
    AdaIR \cite{adair} & 17.70 & 0.261 & 14.15 & 0.375 \\
    PromptIR \cite{potlapalli2024promptir} & 16.78 & 0.21 & 10.94 & 0.431 \\
    AwRaCle \cite{rajagopalan2024awracle} & 16.47 & 0.26 & 17.22 & 0.328 \\
    \rowcolor{YellowGreen!40}
    \textbf{UnSCAR} & \textbf{21.96} & \textbf{0.12} & \textbf{20.59} & \textbf{0.255} \\
    \hline
\end{tabular}
        }
    \end{minipage}
\end{table}
\begin{figure}[t]
    \centering
    \includegraphics[width=.87\linewidth]{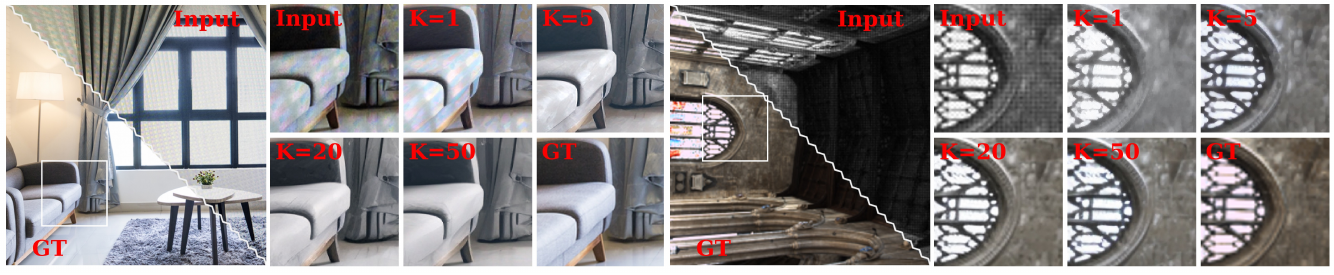}
    \caption{\textit{Fewshot generalization.} One (K=1) to Few (K<=50) shot generalization performance over demoiring and demosaicing tasks.}
    \label{fig:fewshot}
\end{figure}

    
    
    

\vspace{-0.05cm}
\begin{table}[t]
    \centering
    \begin{minipage}[t]{0.49\textwidth}
        \centering
        \caption{\textit{OOD Adaptation.} Generalization to unseen medical image restoration.}
        \label{tab:ood}
        \vspace{-0.25cm} 
        
        \resizebox{\textwidth}{!}{ 
            
  \begin{tabular}{l | ccc | ccc}
    \hline
    \multirow{2}{*}{\textbf{Method}} & \multicolumn{3}{c|}{\textbf{Fundus IR \cite{fundus}}} & \multicolumn{3}{c}{\textbf{Laparoscopic IR \cite{laparoscopic}}} \\
    \cline{2-7} 
    & PSNR$\uparrow$ & SSIM$\uparrow$ & DISTS$\downarrow$ & PSNR$\uparrow$ & SSIM$\uparrow$ & DISTS$\downarrow$ \\ 
    \hline
    
    $\text{BioIR}_1$ \cite{bioIR} & 31.94 & 0.96 & 0.09 & 28.55 & 0.90 & 0.07 \\
    
    $\text{BioIR}_2$ \cite{bioIR} & 31.23 & 0.93 & 0.08 & 27.10 & 0.82 & 0.06  \\
    \rowcolor{YellowGreen!40}
    \rowcolor{YellowGreen!40}
    \textbf{UnSCAR} & 29.60 & 0.95 & 0.09 & 27.81 & 0.95 & 0.05 \\
    \hline
  \end{tabular}
  
        }
    \end{minipage}
    \hfill 
    \begin{minipage}[t]{0.47\textwidth}
        \centering
        \caption{\textit{Few-shot Adaptation} to unseen tasks: Demoiréing and Demosaicing.}
        \label{tab:fewshot}
        \vspace{-0.25cm} 
        
        \resizebox{\textwidth}{!}{ 
           \begin{tabular}{l | ccc | ccc}
    \hline
    \multirow{2}{*}{\textbf{Support Set}} & \multicolumn{3}{c|}{\textbf{Demoiering \cite{yu2022towards}}} & \multicolumn{3}{c}{\textbf{Demosaicing \cite{MSRDemosaicing}}} \\
    \cline{2-7} 
    & PSNR$\uparrow$ & SSIM$\uparrow$ & LPIPS$\downarrow$ & PSNR$\uparrow$ & SSIM$\uparrow$ & LPIPS$\downarrow$ \\
    \hline
    
    $K=1$ & 20.67 & 0.76 & 0.24 & 20.09 & 0.77 & 0.291 \\

    $K=5$ & 21.85 & 0.82 & 0.18 & 22.09 & 0.83 & 0.213 \\
    
    $K=20$ & 22.37 & 0.84 & 0.17 & 22.18 & 0.84 & 0.210 \\

    $K=50$ & 22.58 & 0.89 & 0.15 & 22.33 & 0.89 & 0.181 \\

    \hline
  \end{tabular}
        }
    \end{minipage}

\end{table}

    
    
    

\begin{table}[t]
    \centering
    
    \begin{minipage}[t]{0.4\textwidth}
        \centering
       \caption{\textit{In-the-wild performance} against DA-CLIP~\cite{luo2023controlling} variants.}
        \label{tab:itw}
        \vspace{-2mm} 
        \scriptsize
        \resizebox{\textwidth}{!}{

\begin{tabular}{l | ccc}
    \hline
    \multirow{2}{*}{\textbf{Method}} & \multicolumn{3}{c}{\textbf{Real-Fog} \cite{REVIDE}} \\
    \cline{2-4} 
    & PSNR$\uparrow$ & SSIM$\uparrow$ & LPIPS$\downarrow$ \\
    \hline
    
    $\text{DACLIP}_{\text{UIR}}$  & 14.23 & 0.751 & 0.428 \\
    
    $\text{DACLIP}_{\text{mix}}$ & 15.44 & 0.66 & 0.433 \\

    $\text{DACLIP}_{\text{W}}$ & 14.18 & 0.61 & 0.471  \\
    
    \rowcolor{YellowGreen!40}
    \textbf{UnSCAR} & 16.07 & 0.72 & 0.353  \\
    \hline
  \end{tabular}
            }
    \end{minipage}
    \hfill 
    \begin{minipage}[t]{0.55\textwidth}
        \centering
        \caption{\textit{Ablation studies.} Impact of sampling steps, low-level cues, and control signal strength.}
        \label{tab:abla1}
        \vspace{-2mm} 
        
        \resizebox{\textwidth}{!}{

  \begin{tabular}{l | cc | cc | cc}
    \hline
    \multirow{2}{*}{\textbf{\%age}} & \multicolumn{2}{c|}{\textbf{Sampling Steps} } & \multicolumn{2}{c|}{\textbf{Low-level cues}} & \multicolumn{2}{c}{\textbf{Control Strength}} \\
    \cline{2-7} 
    & PSNR$\uparrow$ & LPIPS $\downarrow$ & PSNR $\uparrow$ & LPIPS $\downarrow$ & PSNR $\uparrow$ & LPIPS$\downarrow$ \\
    \hline

    $\% = 1.0$ & 28.29 & 0.126 & 28.29 & 0.126 & 28.29 & 0.126 \\
    $\% = 0.5$ & 28.48 & 0.131 & 28.20 & 0.131 & 24.52 & 0.254 \\
    $\% = 0.2$ & 27.80 & 0.132 & 27.61 & 0.136 & 19.84 & 0.438 \\
    $\% = 0.1$ & 27.88 & 0.144 & 26.96 & 0.144 & 16.21 & 0.615 \\
    
    \hline
  \end{tabular}
  
        }
    \end{minipage}

\end{table}
We evaluate {\modelName} on seen and unseen mixed-degradation datasets in a zero-shot setting (\cref{tab:unseen_mixed}). {\modelName} consistently outperforms prior approaches by significant margins on both reference and no-reference metrics. Qualitative results (\cref{fig:unseen,fig:mixed}) confirm superior restoration of fine structures and deblurred scenes, particularly on severely degraded cases such as low-light POLED and lowlight+blur images. On the in-the-wild dehazing dataset REVIDE~\cite{REVIDE}, {\modelName} outperforms DA-CLIP~\cite{luo2023controlling} on PSNR and LPIPS while remaining competitive on SSIM (\cref{tab:itw,fig:itw}), demonstrating strong generalization to compound degradations across both in-domain and out-of-domain scenarios.


\vspace{-0.3cm}
\subsection{Additional Experiments}
\begin{figure}[h!]
    \centering
    \includegraphics[width=.9\linewidth]{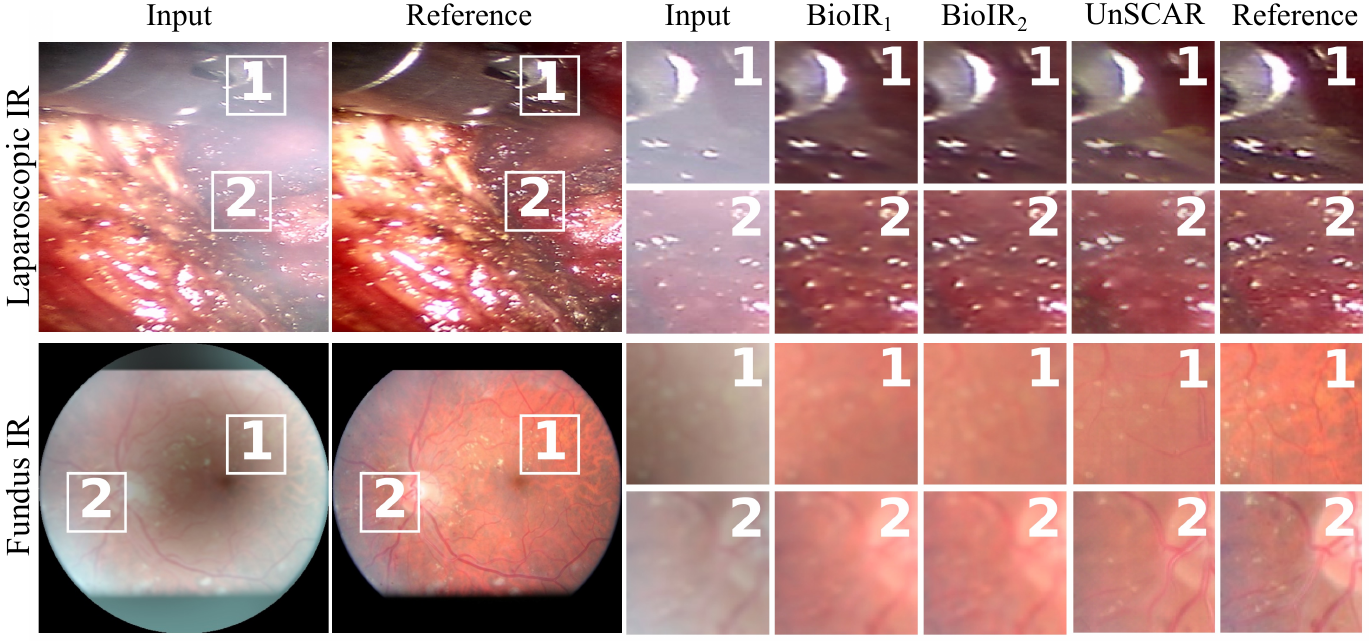}
    \caption{\textit{Medical.} Our model adapts to out-of-distribution data such as medical images.}
    \label{fig:medical}
\end{figure}

\vspace{-0.1cm}
\subsubsection{OOD and Few-shot Task Adaptation:} 
\label{sec:comp_ood} 
Taking a step further, we extend {\modelName} to completely out-of-distribution tasks -- medical image restoration. Specifically, we test on Laparoscopic De-smoking \cite{laparoscopic} and Fundus Image Restoration \cite{fundus}. As described in \cref{sec:comp_ood}, we employed our PEFT method to fine-tune {\modelName} on these benchmarks, and compared against a recent SOTA method BioIR \cite{bioIR} in \cref{tab:ood}, with qualitative samples in \cref{fig:medical}. {\modelName} shows highly competitive performance against the specialist in-domain BioIR (single$_1$- and multi$_2$-task) across all metrics ($\Delta_{PSNR}\approx2.0\downarrow$, $\Delta_{SSIM}\approx0.02\uparrow$, $\Delta_{DISTS}\approx0.01\downarrow$), showing robust generalization to OOD settings. Additionally, we also tested few-shot adaptability on two additional tasks -- Demoiring \cite{yu2022towards} and Demosaicing \cite{MSRDemosaicing} -- where we varied the few-shot support set size, tabulated in \cref{tab:fewshot} (visual samples in \cref{fig:fewshot}). {\modelName} improves with increased support samples as well as show reliable performance even under low-data regimes.



\vspace{-0.3cm}
\subsubsection{Test-time Controllability via Sliders:}
\label{sec:controlable} 
\begin{figure}[t]
    \centering
    \includegraphics[width=0.95\linewidth]{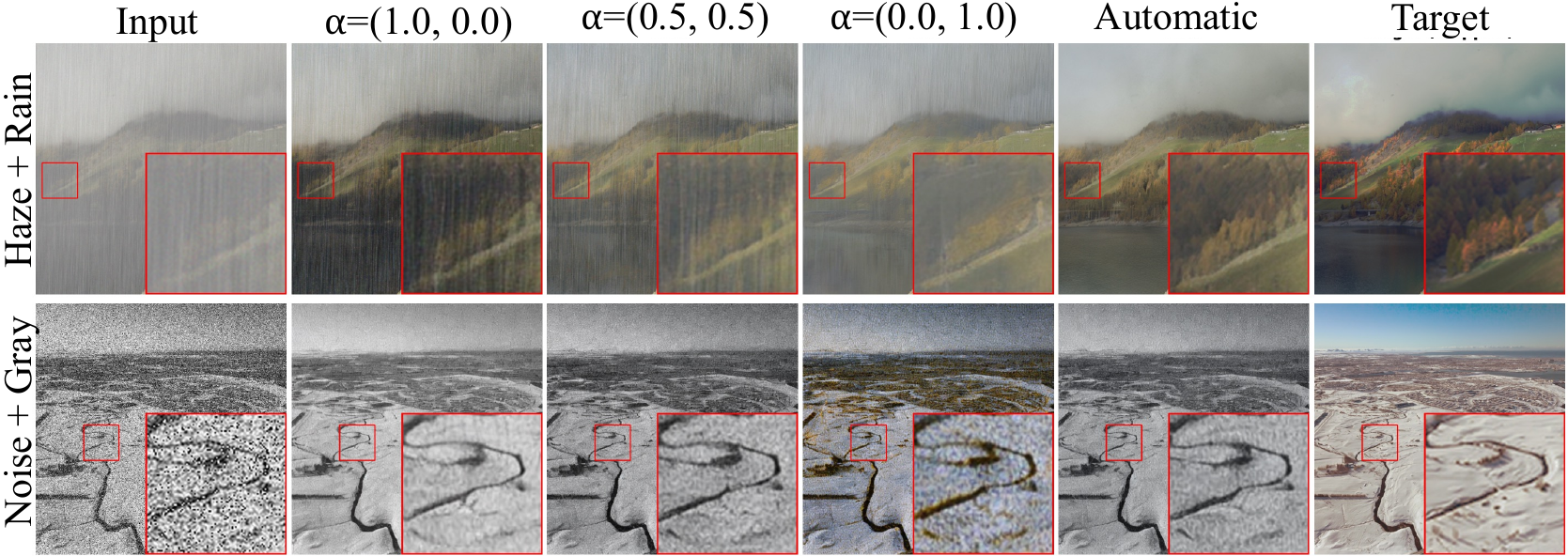}
    \caption{\textit{Controllability.} Our model supports slider-based control of the level of restoration per degradation. $\alpha$ denotes the aberration strength (haze vs. rain, noise vs. gray).}
    \label{fig:control}
\end{figure}

Lastly, we examine {\modelName}'s test-time slider control (described previously in \cref{sec:controlable}), showing qualitative results in \cref{fig:control}. the $\alpha$ values denote the respective weights given to the degradations present. We can see that the model faithfully follows the weight assigned to each degradation i.e. if dehazing is higher it prioritizes haze removal, and so on. This asserts the reliability of our control sliders during test-time for {\modelName}.




\begin{figure}[h!]
    \centering
    \includegraphics[width=0.93\linewidth]{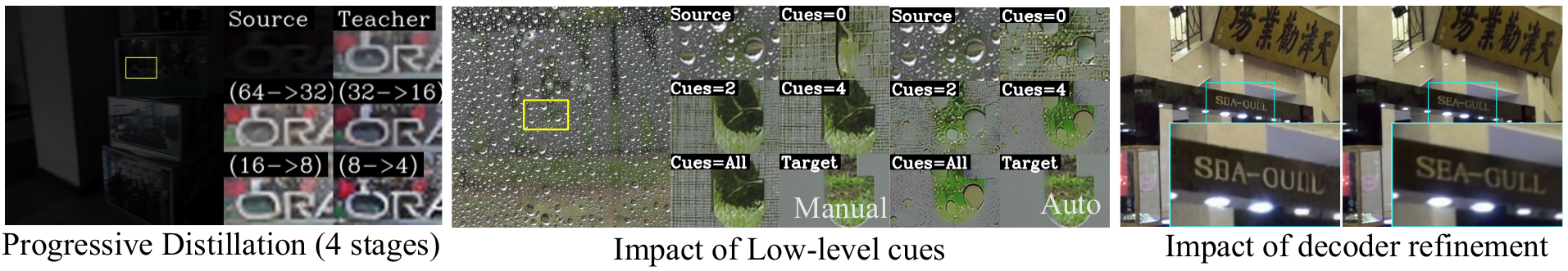}
    \caption{\textit{Ablation studies.} We analyze model performance under reduced sampling steps (left) and fewer low-level cues under automatic versus user controlled restoration (middle), and the impact of decoder refinement on structural preservation (right).}
    \label{fig:abla}
\end{figure}

\vspace{-3mm}
\subsection{Ablation Study} 
\vspace{-0.5mm}

We perform ablations on our model to test restoration quality with respect to minimum number of sampling steps required for inference, impact of low level cues ($v_i$), control signal strength (default is 1.0) and impact of decoder finetuning as shown in \cref{fig:abla} and \cref{tab:abla1}. We perform progressive distillation \cite{meng2023distillation} of our frozen diffusion backbone (keeping restoration control module frozen) to reduce sampling to 4 steps ($\sim$ 10x) without loss of quality. Loss of low-level cues impact the quality of restoration heavily during automatic restoration when no user prior knowledge is available about existing degradations. Decoder finetuning helps preserves structural details such as text lost during the diffusion process.

\vspace{-0.25cm}

\section{Conclusion}
We presented \modelName, a universal image restoration framework designed to scale across diverse and simultaneous degradation types while enabling controllable and adaptable restoration.
Our method combines degradation-aware representations with a mixture-of-experts generative backbone, mitigating cross-degradation interference and supporting stable learning across a large set of degradations.
In addition, we proposed degradation control sliders as a simple yet effective mechanism for user-driven restoration. 
We also introduced a lightweight parameter-efficient adaptation strategy, enabling efficient transfer to unseen domains.
Extensive experiments across single, mixed, and out-of-distribution restoration benchmarks demonstrate the superiority, robustness and flexibility of the proposed approach.
\modelName represents a step toward practical, general-purpose restoration systems capable of operating reliably under diverse real-world conditions.
We hope this work motivates future research toward scalable and controllable frameworks for universal image restoration.


%

\newpage

\bibliographystyle{splncs04}
\bibliography{main}
\end{document}


\title{\textsc{UnSCAR}: \underline{Un}iversal, \underline{S}calable, \underline{C}ontrollable, and \underline{A}daptable Image \underline{R}estoration} 

\titlerunning{Abbreviated paper title}

\author{First Author\inst{1}\orcidlink{0000-1111-2222-3333} \and
Second Author\inst{2,3}\orcidlink{1111-2222-3333-4444} \and
Third Author\inst{3}\orcidlink{2222--3333-4444-5555}}

\authorrunning{F.~Author et al.}

\institute{Princeton University, Princeton NJ 08544, USA \and
Springer Heidelberg, Tiergartenstr.~17, 69121 Heidelberg, Germany
\email{lncs@springer.com}\\
\url{http://www.springer.com/gp/computer-science/lncs} \and
ABC Institute, Rupert-Karls-University Heidelberg, Heidelberg, Germany\\
\email{\{abc,lncs\}@uni-heidelberg.de}}

\maketitle
\section{Supplementary}
\subsection{Model Architecture}
We present our \modelName model's control flow in details in Algorithm 1.a. Specifically, we describe the foundational work \textsc{UniCoRN} first upon which our model builds and improves upon significantly. Next, we highlight the key modifications in red in Algorithm 1.b which differentiates our work from the prior work in terms of architectural novelty. 

\paragraph{\textit{Decoder finetuner module.}} 

\subsection{Implementation details}

\subsection{Experiments discussions}

\subsection{Ablation Studies}

\subsection{More Visual Results}

\bibliographystyle{splncs04}
\bibliography{main}